\documentclass[lettersize,journal]{IEEEtran}
\usepackage{amsmath,amsfonts}
\usepackage{algorithmic}
\usepackage{array}
\usepackage{booktabs}
\usepackage{textcomp}
\usepackage{stfloats}
\usepackage{url}
\usepackage{diagbox}
\usepackage{multirow}
\usepackage{amsmath}
\usepackage{verbatim}
\usepackage{algorithm, algorithmic}
\usepackage{graphicx}
\usepackage{subfigure}
\def\BibTeX{{\rm B\kern-.05em{\sc i\kern-.025em b}\kern-.08em
    T\kern-.1667em\lower.7ex\hbox{E}\kern-.125emX}}
\usepackage{balance}
\newcommand{\myvspaceAbove}{-0.1cm}
\newcommand{\myvspaceAfter}{-0.2cm}

\begin{document}
\title{Boundary Exploration for Next Best View Policy in 3D Robotic Scanning}
\author{Leihui Li, Lixuepiao Wan, Xuping Zhang
\thanks{All authors are with the Department of Mechanical and Production Engineering, Aarhus University, Aarhus, Denmark\par
		$^*$Corresponding author: Xuping Zhang, Email: xuzh@mpe.au.dk}}

\markboth{\;}%
{\;}

\maketitle

\begin{abstract}

The Next Best View (NBV) problem is a pivotal challenge in 3D robotic scanning, with the potential to significantly improve the efficiency of object capture and reconstruction. Existing methods for determining the NBV often overlook view overlap, assume a fixed virtual origin for the camera, and rely on voxel-based representations of 3D data. To address these limitations and enhance the practicality of scanning unknown objects, we propose an NBV policy in which the next view explores the boundary of the scanned point cloud, with overlap intrinsically considered. The scanning or working distance of the camera is user-defined and remains flexible throughout the process. To this end, we first introduce a model-based approach in which candidate views are iteratively proposed based on a reference model. Scores are computed using a carefully designed strategy that accounts for both view overlap and convergence. In addition, we propose a learning-based method, the Boundary Exploration NBV Network (BENBV-Net), which predicts the NBV directly from the scanned data without requiring a reference model. BENBV-Net estimates scores for candidate boundaries, selecting the one with the highest score as the target for the next best view. It offers a significant improvement in NBV generation speed while maintaining the performance level of the model-based approach. We evaluate both methods on the ShapeNet, ModelNet, and 3D Repository datasets. Experimental results demonstrate that our approach outperforms existing methods in terms of scanning efficiency, final coverage, and overlap stability, all of which are critical for practical 3D scanning applications. The related code is available at \url{github.com/leihui6/BENBV}.

\end{abstract}


\section{Introduction}
The ability to efficiently and automatically scan and reconstruct 3D objects or environments, often referred to as view planning, sensor planning, or active perception, is essential in various applications, including industrial inspection \cite{alarcon2014viewpoint}, cultural heritage preservation \cite{giakoumidis2024arm4ch}, and autonomous robotics \cite{batinovic2022shadowcasting}. 
\begin{figure}[htbp]
	\centering
	\vspace{\myvspaceAbove}
	\setlength{\abovecaptionskip}{\myvspaceAbove}
	\includegraphics[width=1.0\linewidth]{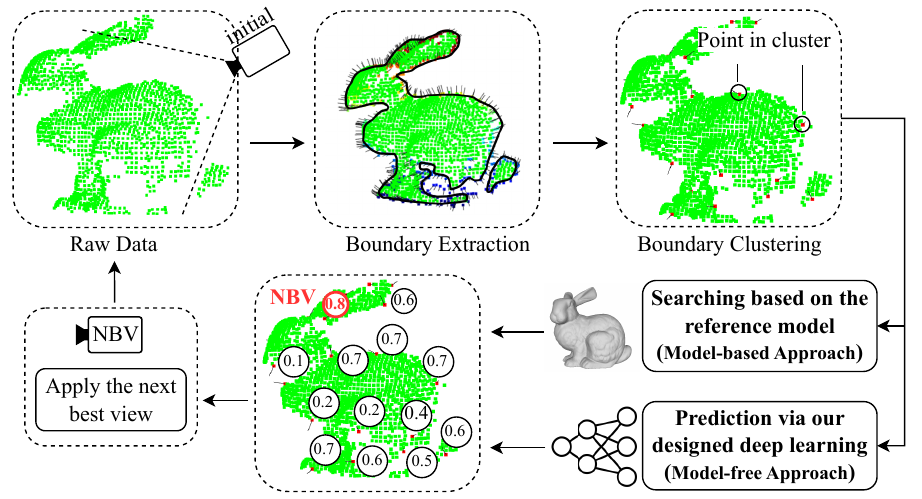}
	\caption{Overview of the proposed approach. Starting from an initial scan, the boundary of the raw point cloud is detected and clustered to generate candidate views. In the model-based approach, the NBV is selected through a search on the reference model. In the model-free approach, scores are predicted for each boundary, and the highest-scoring view is chosen as the NBV.}
	\label{fig.framework}
	\vspace{\myvspaceAbove}
\end{figure}
A key challenge is the Next Best View (NBV) problem, which seeks to determine the optimal sequence of views that maximizes the completeness and quality of a 3D scan while minimizing the number of scans required. An efficient NBV policy enables robotic systems, integrated with vision systems, to maximize the coverage of objects or environments with minimal scanning steps, ensuring high-quality 3D reconstructions and benefiting downstream applications. Several constraints are considered in the NBV policy, including object occlusion, unseen region prediction, sensor movement costs, potential offsets, field of view, and resolution of sensors. The NBV problem, proven to be NP-complete when relying on prior geometric knowledge \cite{tarbox1995planning}, becomes even more challenging when addressing unknown objects or environments, where geometric information is often sparse or incomplete.

The NBV problem has been widely explored using 2D images, but 2D approaches are often limited by the lack of depth information and ambiguity in spatial relationships. In contrast, depth data or point clouds inherently capture rich and geometric details about the target and scene, making them more effective for NBV planning in complex environments. NBV methods utilizing 3D data from point clouds can broadly be categorized into model-based and model-free approaches \cite{lee2024autonomous}. Model-based methods leverage pre-built or provided reference models of target objects to predict the next best view, with the reference model involved in each step. In contrast, model-free approaches generate the NBV without prior knowledge of the object's reference model, relying solely on the information acquired during the scanning process. In addition, most existing NBV methods utilize mesh or voxel representations of the retrieved 3D point cloud, which are well-suited for large outdoor scenes with low precision requirements. However, these methods require additional preprocessing, such as surface reconstruction and  voxelization. These processes increase computational and storage costs while offering lower precision for indoor environments or small objects.

On the other hand, existing methods typically rely on empirically designed view spaces, such as a hemisphere around the target object. However, these approaches limit their generalizability to novel objects. Additionally, they overlook the optimization of the sensor’s working distance, assuming the origin point is at the center of the hemisphere, which can lead to variations in imaging quality. Furthermore, while many existing methods assess scanning efficiency based on coverage, they often overlook the significance of overlap between newly acquired data and existing datasets. This overlap is crucial for properly aligning new data with the existing dataset, especially during the initial scanning stage or when the camera deviates from its expected position and orientation.

To address these challenges and improve NBV selection performance, we introduce a boundary exploration NBV policy. This approach identifies candidate views from the boundary points of the raw accumulated point cloud, directing the camera’s focus toward these boundaries. Specifically, we first propose a model-based method that iteratively searches for and selects the NBV with the highest score. This score balances overlap and coverage to optimize scanning efficiency. Second, we present a model-free method based on a deep learning framework, the Boundary Exploration NBV Network (BENBV-Net). BENBV-Net predicts scores for the boundary points of the point cloud, and the point with the highest score is selected as the NBV. This model eliminates the need for a reference model while significantly improving the inference speed of NBV selection.

In summary, the contributions of this work are summarized as follows:

\begin{enumerate} 
    \item We propose an intuitive NBV policy that directly explores the boundaries of point clouds using raw input data. The approach considers both view overlap and coverage, ensuring stable overlap and high efficiency during data acquisition.
    \item We introduce a model-based approach in which a reference model is used to iteratively select the view with the highest score as the NBV. In addition, we present a learning-based framework, BENBV-Net, which eliminates the need for a reference model by taking the scanned point cloud and its boundaries as input.
    \item The efficiency and effectiveness of our methods are evaluated against traditional approaches on public datasets. BENBV-Net demonstrates strong generalization to unseen data and novel objects, achieving performance comparable to the model-based method while significantly reducing computational time.
\end{enumerate}

\section{Related Work}

Given 3D data, researchers have explored various approaches to predict the optimal next view for capturing target objects. One such approach \cite{kriegel2015efficient, morreale2019predicting} involves using mesh representations of the 3D point cloud, where the mesh is utilized by the robot to identify unknown or poorly reconstructed regions. However, this approach incurs additional computational and storage costs beyond the view prediction itself. An alternative strategy uses volumetric representation \cite{aravecchia2023next, yu2024semantic, naazare2022online}, which divides the scene into multiple voxels, each associated with its observation status. This approach is commonly applied in building octomaps \cite{hornung2013octomap, batinovic2022shadowcasting} or environmental maps, making it effective for capturing full 3D structures and handling occlusions. However, it comes with trade-offs: volumetric methods are computationally and memory-intensive and may compromise surface detail, especially at lower grid resolutions. To address these limitations, our approach directly processes the raw point cloud, allowing for more accurate and efficient perception of small objects and indoor environments, where high resolution and fine surface details are crucial.

Additionally, the NBV problem in model-free scenarios is particularly challenging due to the lack of prior knowledge about the target object. The difficulty lies in selecting viewpoints that efficiently capture the scene \cite{border2024surface}, especially when considering time constraints and unexpected occlusions. To address these challenges, various methods have been proposed. For example, the Surface Edge Explorer (SEE) \cite{border2024surface, border2018surface, border2020proactive} efficiently selects views to improve surface coverage while minimizing movement time. However, it depends on user-defined parameters and struggles with complex geometries. Prediction-guided methods, such as Pred-NBV \cite{dhami2023pred}, maximize information gain by intelligently navigating around obstacles, but may face difficulties in dynamic environments. Learning-based methods like PC-NBV \cite{zeng2020pc} and NBV-net \cite{mendoza2020supervised} introduce deep neural networks that directly process raw point cloud data and current view selection states to predict the information gain of candidate views. However, they are limited by the view space and sensor working distance. Reinforcement learning techniques, such as GenNBV \cite{chen2024gennbv} and RL-NBV \cite{wang2024rl}, adaptively explore environments and achieve high coverage ratios, but they require extensive training data, and their selection policies are often not explainable, making the underlying principles unclear.

It is important to note that when minimizing the number of scan views, it is crucial to account for the overlaps between the newly acquired data and existing views \cite{mendoza2020supervised, pito1999solution}. Failing to do so can make it more difficult to accurately register the views \cite{gazani2023bag}, potentially leading to errors in the reconstruction process. Additionally, the optimal working distance for the 3D camera must be considered or adjusted flexibly to ensure the best imaging performance and high-quality 3D point clouds. 

The proposed NBV policy extends the accumulated data by exploring the boundary of the point cloud. Overlap and coverage are jointly considered through a thoughtfully designed strategy. Additionally, our method allows flexible adjustment of the distance between the target surface and the 3D acquisition device, optimizing the scanning process without assuming the object’s centroid or a fixed origin as the camera’s focal point.

\section{Problem Definition and Methodology}

\subsection{Problem  Definition}

The Next Best View (NBV) problem in 3D robotic scanning involves determining the next optimal viewpoint to capture data about an object or scene, aiming to enhance the completeness and accuracy of 3D scanning. In our study, the surface data of the object, \( s_{i} \in \mathbb{R}^{3} \) perceived by the 3D sensor is represented as a point cloud. The view information and the optimization variables are defined as \( (V^{cam}_{i}, V^{tar}_{i}) \), where \( V^{cam}_{i} \in \mathbb{R}^{3} \) represents the camera's position, and \( V^{tar}_{i} \in \mathbb{R}^{3} \) denotes the focal point of the camera. Notably, \( V^{tar}_{i} \) is always located on the boundary points of the point cloud.

Given an initial point cloud $s_{i}$, the objective of this paper is to determine $(V^{cam}_{i}, V^{tar}_{i})$ to accelerate the data acquisition process while maximizing the objective function $F$, as

\begin{equation}
    \label{eq.pro_def}
    \mathop{\rm argmax} \limits_{V_i \in V} F\mathopen( s_i \left( V^{cam}_{i}, V^{tar}_{i} \right) \mathclose)
\end{equation}

The metric $F$ is defined based on coverage and overlap ratios. Specifically, the coverage is calculated as $\text{C} = \lvert s_i \rvert / \lvert S \rvert$, where $S$ denotes the total number of points in the reference model, and $s_i$ represents the number of points in the retrieved point cloud. The overlap ratio is given by $\text{O} = \lvert p_o \rvert / \lvert s_i \rvert$, where $p_o$ is the number of overlapping points between the newly scanned data and the accumulated point cloud. 

In a simulated environment, the target object is predefined, allowing $S$ to be known, and $s_i$ can be generated using a physics-based simulation engine. In contrast, for real-world applications, $s_i$ is obtained from actual 3D sensor data, while $S$ is typically unknown due to the novelty of the target object.

\subsection{Framework Overview}

Given the captured point cloud, the general idea of this work is to explore unseen regions by following its boundary. However, not all boundary regions contribute meaningfully to achieving high coverage and overlap, and are therefore not always worth exploring or extending. To address this, we first introduce a model-based approach that utilizes a reference model. In this approach, the next-best view (NBV) is determined through a search process, and the view with the highest score is selected as the optimal NBV.

We then propose a model-free approach, in which a deep learning network is trained to predict the view selection score without relying on a reference model. The network is trained on a dataset generated by the model-based approach and learns a latent space mapping between the designated boundary and the corresponding point cloud. The overall framework is illustrated in Figure. \ref{fig.framework}.

\subsection{Boundary exploration for NBV policy}
The boundaries of retrieved 3D representation data have been examined in previous studies \cite{kriegel2011surface, border2024surface}, which utilize either triangle surfaces or density-based methods to explore edges. In our study, we compute the boundary directly from the raw point cloud data using the Angle Criterion method proposed by \cite{bendels2006detecting} where the angle threshold is set 120 degree. Let $B_i$ represent the boundary points for the $i$-th scanned data. To reduce the large number of boundary points and simplify the process, we use the K-Means algorithm to cluster the boundary points into 20 clusters. Within each cluster, the point closest to the cluster centroid is selected and denoted as $V^{\text{tar}}_{i}$. In addition, the set of $V^{\text{tar}}_{i}$ is resorted using a greedy nearest neighbor traversal, where each subsequent point is selected as the closest unvisited one to the current point.

\begin{figure}[htbp]
	\centering
	\vspace{\myvspaceAbove}
	\setlength{\abovecaptionskip}{\myvspaceAbove}
	\includegraphics[width=1.0\linewidth]{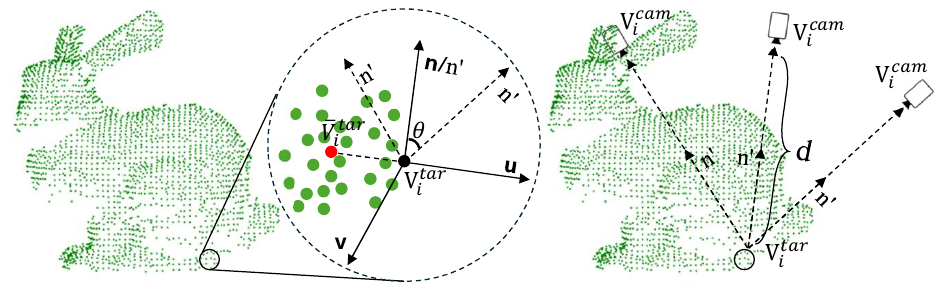}
	\caption{$\textbf{u}\text{-}\textbf{v}\text{-}\textbf{n}$: the $V^{tar}_{i}$ is marked in black and the $\bar{V}^{tar}_{i}$ is denoted in red. The camera position is defined on the right where $d$ is the specified distance.}
	\label{fig.uvn}
	\vspace{\myvspaceAbove}
\end{figure}

The normals for the scanned data are estimated as $s_i^{N \times 6}$and are involved to determine the direction of the camera position. More specifically, a local orthogonal coordinate system ($\textbf{u}$-$\textbf{v}$-$\textbf{n}$) is established at $V^{\text{tar}}_{i}$, where $\textbf{n}$ denotes its normal vector, and $\textbf{u}$ is defined by
\begin{equation}
    \textbf{u} = V^{tar}_{i} - \bar{V}^{tar}_{i}
\end{equation}
where $\bar{V}^{tar}_{i}$ represents the centroid of the neighboring points around $V^{tar}_{i}$, and $\textbf{u}$ indicates the direction of exploration originating from $V^{tar}_{i}$. The $\textbf{v}$ is then computed by $\textbf{u} \times \textbf{n}$. The direction of camera position is expressed as
\begin{equation}
    \textbf{n}^{\prime} = R_{\text{v}}(\theta)\cdot \textbf{n}
\end{equation}
where $\theta$ is set to $-45^{\circ}$, $0^{\circ}$, or $45^{\circ}$ in sequence. An angle of $0^{\circ}$ is used to scan the boundary vertically, while $-45^{\circ}$ and $45^{\circ}$ are chosen to address cases where the boundary lies near the edge, which may otherwise result in empty data. Given the scanning distance $d$, which is user-defined and adjustable for each scan, the camera position is determined as follow
\begin{equation}
    V^{cam}_{i} = d \cdot \textbf{n}^{\prime} + V^{tar}_{i}
\end{equation}
as illustrated in Figure \ref{fig.uvn}.

Given the proposed $V^{\text{cam}}_{i}$, $V^{\text{tar}}_{i}$, and the simulated captured data, a score is assigned to each view by evaluating both coverage and overlap. The view with the highest score is selected as the next-best view. The score at the $i$-th view is calculated as
\begin{equation}
    s = (1 - W_c) \cdot O_i + W_c \cdot C_i
\end{equation}
where $O_i$ and $C_i$ represent the estimated overlap and coverage values at the $i$-th view. The weight $W_c$ is defined as 
\begin{equation}
    W_c = \frac{1}{1 + e^{-10 \cdot (C_i -0.5)}}
\end{equation}
where $C_i$ denotes the coverage of the current view. It is assumed that coverage becomes increasingly important, while overlap becomes less significant as the scanning process progresses. Specifically, overlap is weighted more heavily than coverage until the current coverage exceeds 50\%.

Furthermore, since the overlap values fluctuate across different views, a response function $OR(\cdot)$ is designed to smoothly map the overlap values to continuous values as
\begin{equation}
OR(O_i)=
  \left\{
    \begin{array}{c@{\quad}c}
      -6.25 O_i^2 + 5O_i & O_i < 0.4 \\
      1             & 0.4 \leq O_i \leq 0.5 \\
      -4O_i^2 + 4O_i    & O_i > 0.5
    \end{array}
\right.
\end{equation}
where the overlap range between 0.4 and 0.5 is considered the most representative, while values that are too low ($<$ 0.4) or too high ($>$ 0.5) are penalized. The response is zero at both extremes, specifically when $O_i=0$ and $O_i=1$.

The workflow for searching the NBV is detailed in Algorithm. \ref{alg.workflow}. For the initial view, $V^{cam}_{best}$ is chosen from the surface of a sphere with radius $d$, while $V^{tar}_{best}$ is set as the origin point $(0,0,0)$. The scanned data is provided by the simulation engine. The detailed searching process is outlined in Algorithm. \ref{alg.nbv_search}. The dataset generated in Algorithm. \ref{alg.workflow} is constructed for the deep learning network we developed, where the proposed views (viewList), corresponding point clouds (existedData), and the associated scores (SList) are collected.

\begin{algorithm}
\caption{Workflow for NBV Search}
\label{alg.workflow}
\begin{algorithmic}[1]
    \STATE ${\rm P} \gets {\rm ReferencePointCloud}$
    \STATE ${\rm d} \gets 1.2 $
    \STATE ${\rm V^{cam}_{best},\ V^{tar}_{best}} \gets {\rm initialView}(P)$
    \STATE ${\rm TotalScanCount} \gets 15 $
    \STATE ${\rm i} \gets 0 $
    \STATE ${\rm existedData} \gets None $
    \STATE ${\rm Dataset} \gets [\;] $
    \WHILE{$ i < {\rm TotalScanCount}$}
    \STATE ${\rm scannedData} \gets {\rm simulator}(P, V^{cam}_{best},\ V^{tar}_{best}) $
    \STATE ${\rm (V^{cam}_{best},\ V^{tar}_{best}), viewList, SList} \gets {\rm searchNBV}(\rm ... )$
    \STATE ${\rm Dataset} \gets {\rm (existedData, viewList, SList)}$
    \STATE ${\rm existedData} \gets {\rm existedData + scannedData}$
    \ENDWHILE
\end{algorithmic}  
\end{algorithm}
\begin{algorithm}
\caption{SearchNBV}
\label{alg.nbv_search}
\begin{algorithmic}[1]
    \STATE ${\rm \textbf{Input}: P, existedData, d}$
    \STATE ${\rm B} \gets {\rm BoundaryPointsDetection}({\rm existedData}) $
    \STATE ${\{B_i\}} \gets {\rm BoundaryPointsCluster}({\rm {B, K=20}}) $
    \STATE ${\{\left(V^{cam}_{i}, V^{tar}_{i}\right)\}} \gets {\rm GenerateViews}({\rm {B_i, d}}) $
    \STATE ${\rm viewList} \gets {\rm \{{\left(V^{cam}_{i}, V^{tar}_{i}\right)\}}}$
    \STATE ${\rm MaxS} \gets 0;\  {\rm MaxIndex} \gets 0; \; {\rm SList} \gets [] $
    \FOR   {$ \rm \left(V^{cam}_{i}, V^{tar}_{i}\right) \in \rm viewList $}
    \STATE $\rm C, O = simulator(P, existedData, V^{cam}_{i}, V^{tar}_{i})$
    \STATE ${\rm S} \gets W_c^i \cdot C + \left( 1-W_c^i \right) \cdot OR(O)$
    \IF{$\rm S > MaxS$}
        \STATE $\rm MaxS \gets S;\;  SList.add(S)$
        \STATE $\rm MaxIndex \gets i$
    \ENDIF
    \ENDFOR
    \STATE ${\rm \textbf{Return}\; viewList[MaxIndex], viewList, SList}$
\end{algorithmic}  
\end{algorithm}
\vspace{\myvspaceAfter}

\subsection{Deep Learning Architecture}

The goal of the learning-based approach is to predict the NBV based on the detected boundaries, particularly in 3D automatic scanning tasks involving novel objects for which reference models are unavailable. The training dataset includes the point cloud (P), the detected boundaries (B), and their corresponding scores, which are used as prediction targets. To enable the network to efficiently learn to select the best view, the training supervision pair for the BENBV-Net is defined as (P, B, C) where C consists of the density for B and view order. 

We propose BENBV-Net, a deep neural network for hierarchical feature learning and NBV score prediction from 3D point clouds. The network employs two parallel encoders: a point feature encoder for spatial coordinates and a normal feature encoder for surface normals, both utilizing convolutional layers with batch normalization and ReLU activation. The \textbf{P} feature follows a structure similar to PointNet, processing point cloud data in a permutation-invariant manner. Boundary features are extracted via fully connected layers, while contextual features, including point density and view order, are fused using a learnable density and view order fusion module. These features are combined with global representations through residual blocks for multi-scale refinement. A multi-head self-attention mechanism captures long-range dependencies.

\begin{figure*}[htbp]
\centering
\setlength{\abovecaptionskip}{\myvspaceAbove}
\includegraphics[width=1.0\linewidth]{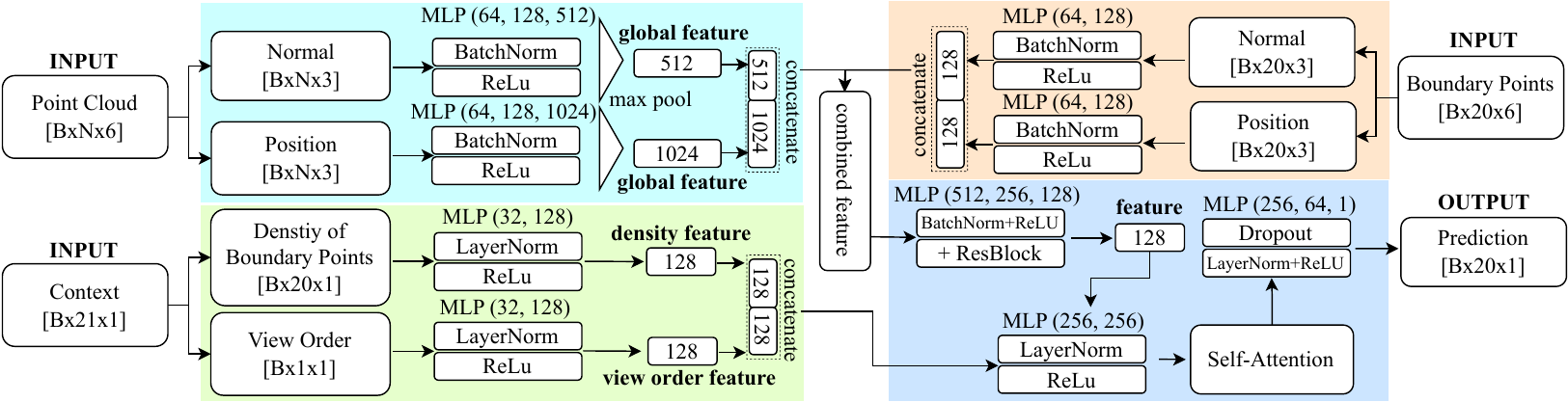}
\caption{Our developed deep learning framework: BENBV-Net takes the raw point cloud, boundary points, and the corresponding context as input. The network first extracts the global feature of the current point cloud, which is then fused with the global feature from the boundary points, along with the features extracted from the context. These combined features are used to predict scores for each boundary point, and the highest-scoring boundary is selected as the NBV.}
\label{fig.dlframework}
\vspace{\myvspaceAbove}
\end{figure*}

The prediction head computes per-point NBV scores using fully connected layers with dropout regularization. The entire network is trained end-to-end from raw point cloud data. An overview is shown in Figure. \ref{fig.dlframework}.

A position-aware loss function for next-best-view prediction is designed which emphasizes predictions at the score of different boundaries. The loss function $L$ for a batch of predictions $y_s^i$ and ground truth values $Y_s^i$ is defined as:

\begin{equation}
    L(y_s, Y_s) = \lambda \sum(w_i(y_s^i - Y_s^i)^{2})
\end{equation}
where $w_i = ((x_i - 12)/10 )^{2} + 0.3$ is the position-dependent weight for the $i$-th view position, $x_i$ is the view order index within the range $\left[0,19\right]$, and $\lambda$ is a scaling factor set to 5.0. This weighting scheme emphasizes predictions at extreme viewing scans, as positions at the beginning are typically more critical and informative for NBV selection than central views. Specifically, the minimum weight is 0.3 for positions near the middle, with front scans receiving higher importance than those in the latter half of the sequence.

\section{Implementation Details and Experiments}
\subsection{Experiment Setup}
We use PyBullet \cite{coumans2021} as the simulation platform, with the camera configuration set to a near distance of 0.01 m, a far distance of 5 m, and a captured image size of $1280 \times 720$ pixels, with a $70^{\circ}$ field of view (FOV). The scanning distance for the 3D camera is set to 1.2 meters, which is an optimal distance for most low-cost cameras. The training of BENBV-Net and all experiments were conducted on a system running Ubuntu 24.04, equipped with an Intel i9 processor and an NVIDIA RTX 3090 GPU.

\subsection{Dataset}

\begin{figure}[htbp]
\centering
\setlength{\abovecaptionskip}{\myvspaceAbove}
\includegraphics[width=1.0\linewidth]{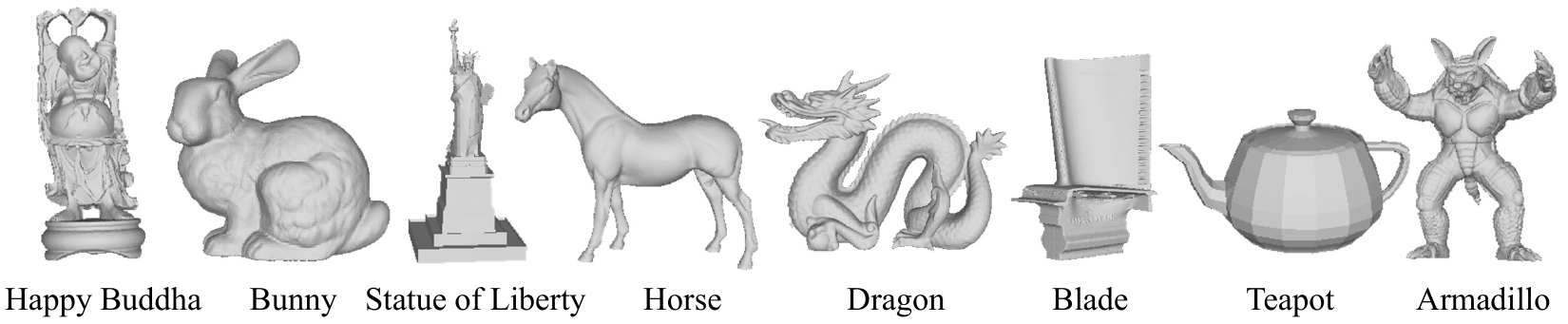}
\caption{An example from the 3D Repository dataset, which contains classic 3D models widely used in the research community. Additionally, point clouds from ShapeNet and ModelNet40 are also included.}
\label{fig.stan3d_dataset}
\vspace{\myvspaceAbove}
\end{figure}

We evaluate and train our model-based method and BENBV-Net using the ShapeNet \cite{shapenet2015}, ModelNet40 \cite{wu20153d} datasets, and 3D Repository from Stanford University \footnote{\url{graphics.stanford.edu/data/3Dscanrep/}} and Georgia Tech \footnote{\url{sites.cc.gatech.edu/projects/large_models}}, with an example shown in Figure \ref{fig.stan3d_dataset}. The ShapeNet and ModelNet40 datasets are used for both training and evaluation, while the 3D Repository is reserved exclusively for evaluation. Specifically, we use ShapeNetV1, selecting 20 models per category for training and 15 models per category for testing. Similarly, for ModelNet40, which includes 40 categories of CAD-generated meshes, we select 20 models per category for training and 15 for testing. For each model, 16 scans are extracted as trainable data. This setup results in a total of over 24,000 training samples from ShapeNetV1 and ModelNet40, 18,000 testing samples, and 128 testing samples from the 3D Repository. The datasets we built and utilized are available online\footnote{\url{huggingface.co/datasets/Leihui/NBV}}. The scanning model used in our simulation platform is sampled as points using Poisson disk sampling \cite{yuksel2015sample}.

\vspace{\myvspaceAfter}


\subsection{Network Training}
Our proposed BENBV-Net is trained using the Adam optimizer with a base learning rate of 0.001 and a mini-batch size of 128. During both training and testing, the input data are randomly downsampled to 4,096 points. The loss converges to 0.002 after approximately 150 epochs. To enhance model generalization, the training dataset is augmented by applying random rotations with angles ranging from -10$^{\circ}$ to 10$^{\circ}$ along the $X$, $Y$, and $Z$ axes. The entire training process takes approximately 2 hours.

\subsection{Evaluation Metrics}

The performance of the comparison methods is evaluated using the following metrics: 1) final coverage, 2) stability of overlap between the newly current view and existing data, 3) Chamfer Distance (CD), 4) Hausdorff Distance (HD), and 5) scanning efficiency. Specifically, we calculate the average of the initial overlap values, as early overlaps are more critical and challenging to optimize in practical scenarios. The coefficient of variation (CV) of the overlap is reported for the first three scans. Furthermore, the scanning quality is evaluated by comparing the final retrieved points with the original model using Chamfer and Hausdorff distances. The Chamfer Distance measures the average distance between two point sets, providing an overall indication of their similarity, while the Hausdorff Distance captures the maximum distance from any point in one set to its nearest neighbor in the other, emphasizing worst-case alignment errors. Additionally, scanning efficiency, which aligns with the objectives of the general next best view policy, is evaluated by
\begin{equation}
	e = \rm c * 100 / \rm v
\end{equation}
where $\rm v$ represents the number of views required to first reach or exceed 90\% coverage, and $\rm c \in [0, 1]$ denotes the final coverage ratio. A higher $e$ value indicates greater efficiency of the NBV policy, meaning that high coverage is achieved with fewer views.

\subsection{Performance Comparison}

We compare our proposed method with several approaches that utilize point clouds as input data, including a learning-based NBV prediction method and a density-based edge exploration method. Specifically, the evaluated methods are:

\begin{enumerate}
\item \textbf{PC-NBV} \cite{zeng2020pc}: A deep neural network for next-best-view planning. The pretrained model provided by the authors is used in this work.
\item \textbf{SEE} \cite{border2024surface}: A surface-edge exploration method that selects views based on point cloud density and edge characteristics. The default parameter settings for small objects are applied in this work.
\item \textbf{Random Boundary}: The next-best view is randomly selected from the detected boundary of point cloud. 
\item \textbf{Random Sphere}: The next camera position is randomly generated on the surface of a sphere.
\item \textbf{Random Uniform Sphere (Random U-Sphere)}: The next camera position is randomly selected from a set of uniformly distributed points on a sphere.
\item \textbf{Ours (BENBV)}: Our method searches for and selects the optimal next view given the detected boundaries.
\item \textbf{Ours via Deep Learning (BENBV-Net)}: Given the detected boundary and captured point cloud, this model directly predicts the next-best view score without performing an explicit search.
\end{enumerate}

Each method is executed 60 times on the 3D Repository dataset and twice on the ShapeNet and ModelNet datasets. For all tested methods, the initial view is randomly selected from a spherical surface. The performance of our NBV policy in comparison with other methods on the ShapeNet, ModelNet, and 3D Repository datasets is summarized in Tables \ref{tab.perf_shapenet}, \ref{tab.perf_modelnet}, and \ref{tab.perf_stanford3d}, respectively. The reported coverage denotes the final coverage achieved after 15 scans. The best results are highlighted in bold, and the second-best results are underlined.

The results show that our method (BENBV) achieves the highest final coverage across all compared approaches. More importantly, it reaches scanning efficiency scores of over 10.0 and 13.0, outperforming all other methods. In addition, our approach consistently selects views with stable overlap, reaching a value of 8.16 on the 3D Repository dataset. The Chamfer and Hausdorff distances, used to evaluate reconstruction quality, are either lower than or on par with those of the other methods, supporting the accuracy of our approach. It is also worth noting that the random boundary method achieves a comparable CV of overlap, suggesting that selecting views along object boundaries is a promising direction for NBV policy design. However, its low efficiency remains a significant drawback.

Regarding BENBV-Net, it generally delivers the best performance among the model-free methods, including PC-NBV and SEE. On the ModelNet40 dataset, PC-NBV achieves lower Chamfer and Hausdorff distances than our method, which is commendable. However, its CV of overlap is the highest (i.e., most unstable), and its scanning efficiency is also lower than that of BENBV-Net. Similarly, on the ShapeNet dataset, PC-NBV again achieves a lower Chamfer distance, but both its efficiency and overlap stability are worse compared to our method. As for SEE, which is based on edge density, it shows comparable overlap stability only on the 3D Repository dataset. On ModelNet and ShapeNet, however, it performs the worst among all evaluated methods. As a result, our proposed methods, BENBV and BENBV-Net, achieve the best and second-best performance in terms of efficiency, final coverage, and overlap stability. Although the final reconstruction quality does not surpass that of PC-NBV, it remains on a comparable level.

\begin{table}[htbp]
\centering
\vspace{\myvspaceAbove}
\caption{Evaluation results of Next-Best-View policies on ShapeNet. 
}
\label{tab.perf_shapenet}
\resizebox{\columnwidth}{!}{%
\large
\begin{tabular}{@{}llllll@{}}
\toprule
                      & \multicolumn{5}{c}{ShapeNet}                               \\ \cmidrule(l){2-6} 
NBV Policy &
  \multicolumn{1}{c}{\begin{tabular}[c]{@{}c@{}}Coverage\\(\%) $\uparrow$\end{tabular}} &
  \multicolumn{1}{c}{\begin{tabular}[c]{@{}c@{}}CV of\\ Overlap $\downarrow$\end{tabular}} &
  \multicolumn{1}{c}{\begin{tabular}[c]{@{}c@{}}CD\\(mm) $\downarrow$\end{tabular}} &
  \multicolumn{1}{c}{\begin{tabular}[c]{@{}c@{}}HD\\(mm) $\downarrow$\end{tabular}} &
  \multicolumn{1}{c}{\begin{tabular}[c]{@{}c@{}}Efficiency\\  $\uparrow$\end{tabular}} \\ \midrule
BENBV	&\textbf{91.27}	&\textbf{12.58}	&\textbf{0.27}	&\textbf{8.62}	&\textbf{9.22}	\\ 
Random Boundary	&86.64	&34.30	&0.38	&10.34	&6.68	\\
Random Sphere	&86.10	&43.80	&0.36	&9.94	&6.42	\\
Random U- Sphere	&86.80	&48.95	&0.35	&10.05	&6.54	\\ 
PC-NBV	&87.36	&50.80	&\underline{0.33}	&9.72	&7.11	\\
SEE	&63.84	&41.46	&1.24	&18.11	&4.20	\\
BENBV-Net	&\underline{88.24}	&\underline{23.19}	&0.35	&\underline{9.62}	&\underline{7.71}	\\ \bottomrule
\end{tabular}%
}
\vspace{\myvspaceAbove}
\end{table}

\begin{table}[htbp]
\centering
\vspace{\myvspaceAbove}
\caption{Evaluation results of NBV policies on ModelNet40.}
\label{tab.perf_modelnet}
\resizebox{\columnwidth}{!}{%
\large
\begin{tabular}{@{}llllll@{}}
\toprule
                      & \multicolumn{5}{c}{ModelNet}                         \\ \cmidrule(l){2-6} 
NBV Policy &
  \multicolumn{1}{c}{\begin{tabular}[c]{@{}c@{}}Coverage\\(\%) $\uparrow$\end{tabular}} &
  \multicolumn{1}{c}{\begin{tabular}[c]{@{}c@{}}CV of\\ Overlap $\downarrow$\end{tabular}} &
  \multicolumn{1}{c}{\begin{tabular}[c]{@{}c@{}}CD\\(mm) $\downarrow$\end{tabular}} &
  \multicolumn{1}{c}{\begin{tabular}[c]{@{}c@{}}HD\\(mm) $\downarrow$\end{tabular}} &
  \multicolumn{1}{c}{\begin{tabular}[c]{@{}c@{}}Efficiency\\  $\uparrow$\end{tabular}} \\ \midrule
BENBV	&\textbf{92.58}	&\textbf{15.33}	&\textbf{0.22}	&\textbf{7.01	}&\textbf{10.16}	\\ 
Random Boundary	&87.78	&33.42	&0.32	&9.18	&7.10	\\
Random Sphere	&88.3	&47.90	&0.27	&8.38	&7.08	\\
Random U- Sphere	&87.93	&49.91	&0.28	&8.32	&6.92	\\ 
PC-NBV	&88.24	&52.41	&\underline{0.27}	&\underline{8.23}	&7.44	\\
SEE	&67.82	&38.86	&1.06	&16.52	&4.61	\\
BENBV-Net	&\underline{89.18}	&\underline{26.18}	&0.31	&8.45	&\underline{8.34}	\\ \bottomrule
\end{tabular}
}
\vspace{\myvspaceAbove}
\end{table}

\begin{table}[htbp]
\centering
\vspace{\myvspaceAbove}
\caption{Evaluation results of NBV policies on 3D repository.}
\label{tab.perf_stanford3d}
\resizebox{\columnwidth}{!}{%
\large
\begin{tabular}{@{}llllll@{}}
\toprule
                      & \multicolumn{5}{c}{3D Repository}                            \\ \cmidrule(l){2-6} 
NBV Policy &
  \multicolumn{1}{c}{\begin{tabular}[c]{@{}c@{}}Coverage\\(\%) $\uparrow$\end{tabular}} &
  \multicolumn{1}{c}{\begin{tabular}[c]{@{}c@{}}CV of\\ Overlap $\downarrow$\end{tabular}} &
  \multicolumn{1}{c}{\begin{tabular}[c]{@{}c@{}}CD\\(mm) $\downarrow$\end{tabular}} &
  \multicolumn{1}{c}{\begin{tabular}[c]{@{}c@{}}HD\\(mm) $\downarrow$\end{tabular}} &
  \multicolumn{1}{c}{\begin{tabular}[c]{@{}c@{}}Efficiency\\  $\uparrow$\end{tabular}} \\ \midrule
BENBV	&\textbf{95.70}	&\textbf{8.16}	&\textbf{0.09}	&\textbf{5.20}	&\textbf{13.45}	\\ 
Random Boundary	&94.28	&23.55	&0.11	&5.91	&9.5	\\
Random Sphere	&91.24	&33.27	&0.15	&6.56	&7.55	\\
Random U- Sphere	&91.97	&34.45	&0.14	&6.47	&7.74	\\ 
PC-NBV	&92.11	&34.76	&0.14	&6.19	&8.76	\\
SEE	&77.67	&27.33	&0.52	&12.34	&5.76	\\
BENBV-Net	&\underline{94.44}	&\underline{18.42}	&\underline{0.11}	&\underline{5.65}	&\underline{10.64}	\\ \bottomrule
\end{tabular}%
}
\vspace{\myvspaceAbove}
\end{table}

To better understand the achieved final coverage, we use the number of views required to reach 50\%, 80\%, and 90\% coverage as key milestones for evaluating NBV policies. These results are detailed in Table \ref{tab.coverage_details}. Random-based methods typically reach 50\% coverage within four scans; however, they require significantly more views to reach higher coverage, around 13 scans for ShapeNet and 10 for the 3D Repository dataset to reach 90\%. In contrast, our BENBV method achieves these milestones more efficiently. For example, it reaches 80\% coverage with approximately 6 views and 90\% with just 7 views on the 3D Repository dataset. BENBV-Net shows similarly strong performance, converging faster than PC-NBV while requiring fewer scans.

It is also worth noting that PC-NBV can reach 50\% coverage with only 3 scans, but this comes at the cost of significantly lower overlap compared to our approach. Since overlap stability is crucial in practical 3D scanning tasks, this limits its overall usefulness. These results highlight a key distinction: while existing methods focus solely on maximizing final coverage, they often overlook the importance of overlap stability. Our method balances both goals: high coverage with fewer scans and stable overlap in the initial scanning phase.

\begin{table}[htbp]
\centering
\vspace{\myvspaceAbove}
\caption{Comparison of NBV policies based on the number of views required to achieve 50\%, 80\%, and 90\% coverage.}
\label{tab.coverage_details}
\resizebox{1.0\columnwidth}{!}{%
\LARGE
\begin{tabular}{@{}llllllllll@{}}
\toprule
\multicolumn{1}{c}{\multirow{2}{*}{NBV Policy}} & \multicolumn{3}{c}{ShapeNet} & \multicolumn{3}{c}{ModelNet40} & \multicolumn{3}{c}{3D Repository} \\ \cmidrule(l){2-10} 
\multicolumn{1}{c}{} & 50\% & 80\% & 90\% & 50\% & 80\% & 90\% & 50\% & 80\% & 90\% \\ \midrule
BENBV	&\textbf{3.94}	&\textbf{7.67}	&\textbf{9.90}	&\underline{3.76}	&\textbf{7.02}	&\textbf{9.11}	&\underline{3.26}	&\textbf{6.30}	&\textbf{7.11}	\\
Random Boundary	&4.54	&10.69	&12.98	&4.24	&9.83	&12.36	&3.78	&8.12	&9.93	\\
Random Sphere	&4.55	&10.81	&13.41	&4.13	&9.82	&12.48	&3.92	&9.13	&12.09	\\
Random U- Sphere	&4.49	&10.48	&13.27	&4.23	&9.97	&12.72	&3.83	&8.74	&11.88	\\
PC-NBV	&\underline{3.53}	&9.14	&12.29	&\textbf{3.47}	&8.87	&11.86	&\textbf{3.22}	&7.49	&10.52	\\
SEE	&6.66	&13.81	&15.21	&6.28	&12.84	&14.71	&4.17	&10.71	&13.47	\\
BENBV-Net	&4.30	&\underline{8.93}	&\underline{11.44}	&4.11	&\underline{8.34}	&\underline{10.70}	&3.42	&\underline{6.94}	&\underline{8.87}	\\ \bottomrule
\end{tabular}%
}
\end{table}

Furthermore, the detailed overlap metrics are presented in Table \ref{tab.overlap_details}, showing the average overlap during the first three scans. In the table, the lowest overlap values, often considered unexpected, are marked with a box for emphasis. According to the results, BENBV consistently achieves stable overlap performance, maintaining around 43\% overlap across the first three scans. BENBV-Net also demonstrates similarly stable overlap, averaging above 40\%.

In contrast, PC-NBV shows some of the lowest overlap values, with averages of 11.92\%, 23.72\%, and 38.13\%, respectively. This helps explain why PC-NBV is able to reach 50\% coverage within just 3 scans (as shown in Table \ref{tab.coverage_details}): it prioritizes coverage while largely sacrificing overlap. In addition, while SEE achieves higher overlap than PC-NBV, Random Sphere, and Random U-Sphere, it still only reaches about half the overlap achieved by our methods, BENBV and BENBV-Net. 

Interestingly, the random boundary method consistently achieves relatively high overlap in the first scan. This suggests that exploring boundaries may be an effective strategy for achieving reliable overlap in NBV planning.

\begin{table}[htbp]
\centering
\vspace{\myvspaceAbove}
\caption{Comparison of NBV policies based on the first 3 scans.}
\label{tab.overlap_details}
\resizebox{0.9\columnwidth}{!}{%
\LARGE
\begin{tabular}{@{}lcccccccccl@{}}
\toprule
\multirow{2}{*}{NBV Policy} & \multicolumn{3}{c}{ShapeNet} & \multicolumn{3}{c}{ModelNet40} & \multicolumn{3}{c}{3D Repository} &  \\ \cmidrule(lr){2-10}
 & 1 $^{th}$ & 2 $^{nd}$ & 3 $^{rh}$ & 1 $^{th}$ & 2 $^{nd}$ & 3 $^{rh}$ & 1 $^{th}$ & 2 $^{nd}$ & 3 $^{rh}$ &  \\ \midrule
BENBV	&43.05	&43.23	&44.68	&43.42	&43.90	&46.30	&43.33	&42.65	&41.88	\\
Random Boundary	&39.42	&51.71	&57.33	&41.37	&50.33	&56.66	&42.01	&52.55	&59.04	\\
Random Sphere	&19.04	&29.73	&39.85	&19.18	&32.54	&39.59	&19.11	&30.78	&40.95	\\
Random U- Sphere	&17.61	&29.05	&\fbox{36.84}	&18.92	&29.45	&\fbox{36.27}	&19.22	&29.55	&39.85	\\
PC-NBV	&\fbox{11.92}	&\fbox{23.72}	&38.13	&\fbox{16.33}	&\fbox{24.70}	&38.21	&\fbox{10.07}	&\fbox{22.21}	&\fbox{35.34}	\\
SEE	&20.21	&51.64	&67.15	&24.00	&53.99	&67.70	&20.23	&46.45	&65.59	\\
BENBV-Net	&40.45	&43.13	&43.40	&40.15	&43.19	&46.55	&43.82	&43.18	&45.04	\\ \bottomrule
\end{tabular}%
}
\vspace{\myvspaceAbove}
\end{table}
\vspace{\myvspaceAfter}

\begin{figure}[htbp]
\centering
\setlength{\abovecaptionskip}{\myvspaceAbove}
\includegraphics[width=0.95\linewidth]{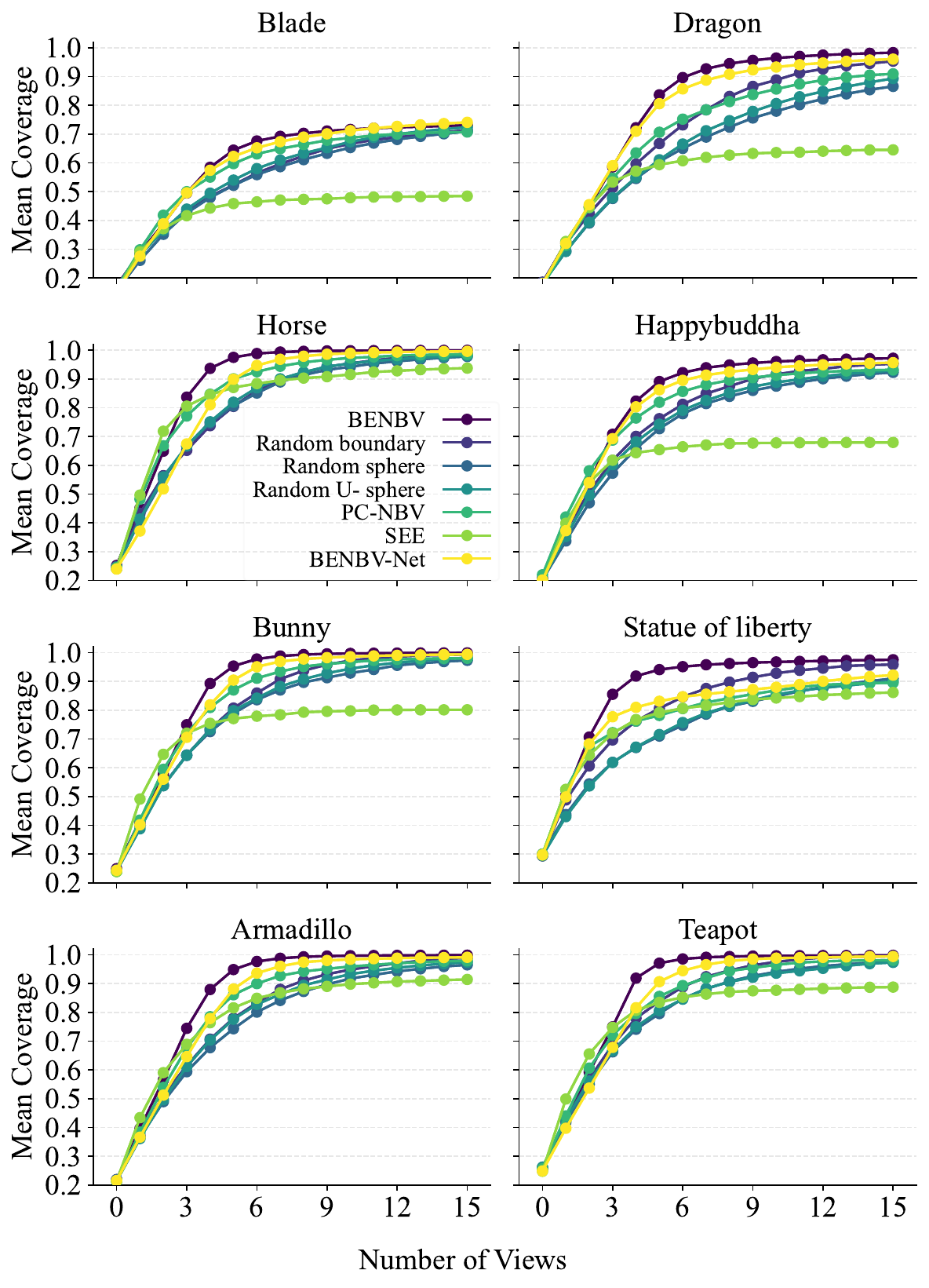}
\caption{Coverage for each view for the 3D Repository dataset.}
\label{fig.stan3d_converge}
\vspace{\myvspaceAbove}
\end{figure}

\begin{figure}[htbp]
\centering
\setlength{\abovecaptionskip}{\myvspaceAbove}
\includegraphics[width=0.95\linewidth]{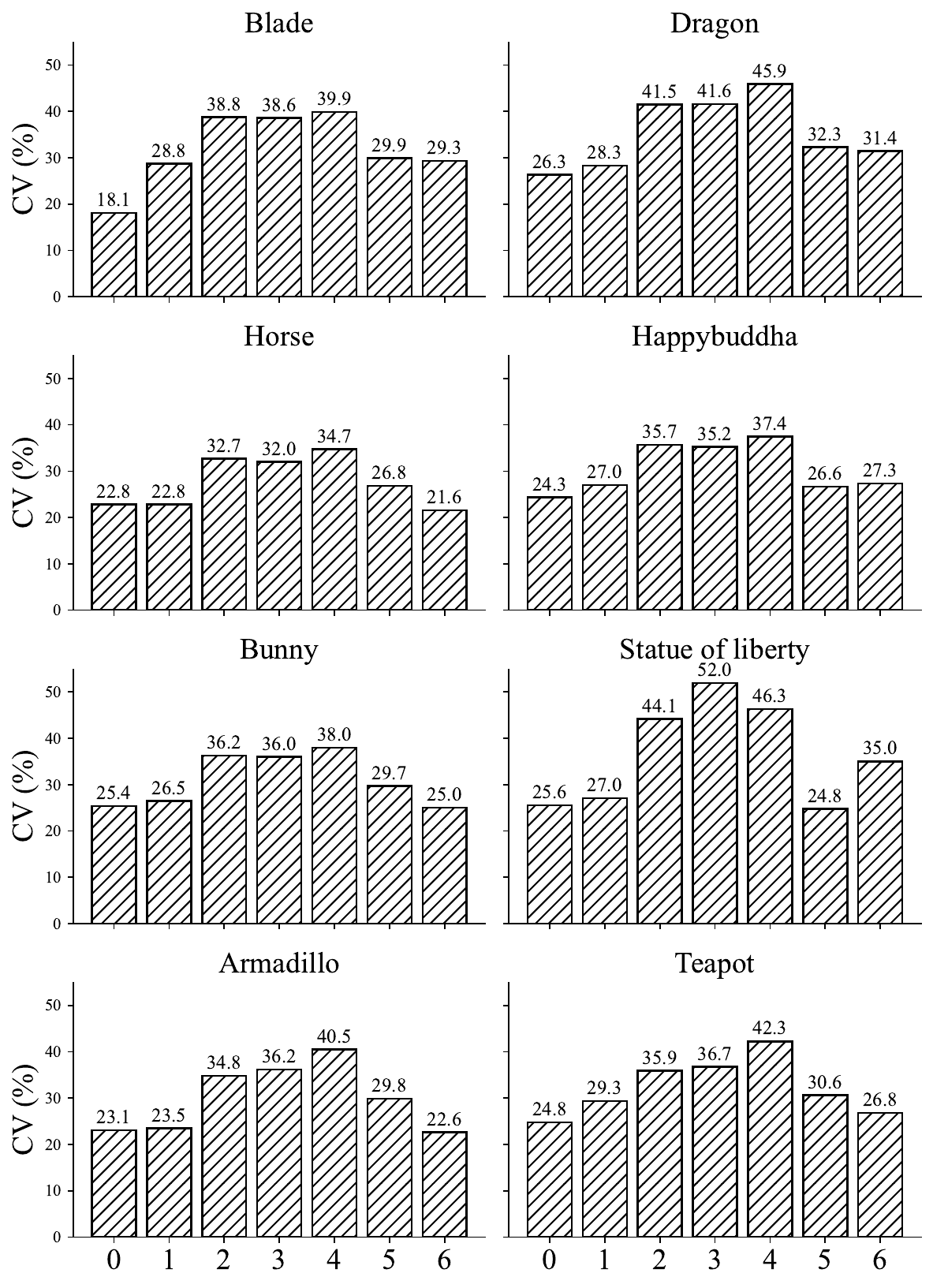}
\caption{CV of overlap over all scans in the 3D Repository Dataset. Methods are labeled as follows: 0: BENBV, 1: Random Boundary, 2: Random Sphere, 3: Random Uniform Sphere, 4: PC-NBV, 5: SEE, 6: BENBV-Net.}
\label{fig.stan3d_overlap}
\vspace{\myvspaceAbove}
\end{figure}

To provide a clearer comparison of the NBV policies, we illustrate the convergence behavior and overlap between views using the 3D Repository dataset as an example, as shown in Figure. \ref{fig.stan3d_converge} and Figure. \ref{fig.stan3d_overlap}. As seen in Figure. \ref{fig.stan3d_converge}, BENBV consistently achieves the highest coverage across most cases. BENBV-Net performs similarly, with the exception of the \textit{Statue of Liberty} sample, where it falls slightly behind. Based on Figure \ref{fig.stan3d_overlap}, both the model-based method and BENBV-Net exhibit stable overlap performance. In some cases, such as the \textit{Bunny} and \textit{Horse} datasets, BENBV-Net performs on par with BENBV. Moreover, since the 3D Repository dataset was not used during training, these results also highlight BENBV-Net’s ability to generalize to unseen objects which is a challenging task even for model-based methods.

A significant drawback of BENBV is its longer processing time, as shown in Table \ref{tab.time_spent}, which reports the average time required to scan each object across the tested datasets. The results indicate that BENBV takes approximately 8 to 13 times longer than other model-free methods. In contrast, our learning-based approach, BENBV-Net, requires only 7.8 seconds per object, demonstrating competitive efficiency in terms of runtime.

\begin{table}[htbp]
	\centering
	\vspace{\myvspaceAbove}
	\caption{The average process time (s)  for each object in tested datasets.}
	\label{tab.time_spent}
	\resizebox{\columnwidth}{!}{%
		\large
		\begin{tabular}{@{}llllllll@{}}
        \toprule
        Dataset & BENBV & \begin{tabular}[c]{@{}l@{}}Random \\ Boundary\end{tabular} & \begin{tabular}[c]{@{}l@{}}Random \\ Sphere\end{tabular} & \begin{tabular}[c]{@{}l@{}}Random \\ U- Sphere\end{tabular} & PC-NBV & SEE & BENBV-Net \\ \midrule
        
        ShapeNetV1  & 63.2 & 7.2 & 4.6 & 4.6 & 5.3 & 6.9 & 7.8 \\
        ModelNet40  & 66.6 & 8.1 & 5.8 & 5.6 & 6.4 & 7.9 & 7.6 \\
        Stanford 3D & 64.0 & 7.0 & 6.9 & 6.2 & 5.9 & 5.3 & 7.8 \\ \bottomrule
		\end{tabular}%
	}
\vspace{\myvspaceAbove}
\end{table}

In summary, the random-based approaches achieve moderate final coverage but require more views to reach 90\% completeness. PC-NBV reaches high coverage quickly but largely overlooks overlap during data acquisition. SEE, on the other hand, ensures high overlap during scanning but struggles to select views that meaningfully increase coverage. Our experiments show that the proposed method achieves higher coverage more efficiently while maintaining stable overlap between views, particularly during the early stages of scanning. In other words, it strikes a balanced trade-off between overlap acquisition and final coverage.



\section{Conclusion and Future Work}

In this paper, we propose an NBV policy that explores the boundary of the point cloud while balancing coverage and overlap. We present two approaches: a model-based method, BENBV, which uses a reference model to search for potential views based on a carefully designed strategy, selecting the view with the highest score as the NBV; and a learning-based method, BENBV-Net, which employs a deep neural network trained on data generated by the model-based approach. BENBV-Net predicts scores directly from the accumulated data, allowing NBV selection without reference models.

The experiments demonstrate that both of our proposed methods achieve high coverage and stable overlap when evaluated on public datasets, including ShapeNet, ModelNet40, and the 3D Repository. Most importantly, the scanning efficiency ranks as the best or second-best among all tested methods. Furthermore, the time efficiency of our learning-based approach is comparable to traditional methods, highlighting its practicality for real-world 3D scanning tasks.

However, the method is not fully end-to-end, as it relies on boundary extraction to predict scores. Additionally, future work will address this limitation by exploring the incorporation of action space for robotic movement.

\section*{Acknowledgments}
The support of Jungner Company is gratefully acknowledged.

\bibliographystyle{unsrt}
\bibliography{refs}

\begin{thebibliography}{10}

\bibitem{alarcon2014viewpoint}
Jose~Luis Alarcon-Herrera, Xiang Chen, and Xuebo Zhang.
\newblock Viewpoint selection for vision systems in industrial inspection.
\newblock In {\em 2014 IEEE International Conference on Robotics and Automation
  (ICRA)}, pages 4934--4939. IEEE, 2014.

\bibitem{giakoumidis2024arm4ch}
Nikolaos Giakoumidis and Christos-Nikolaos Anagnostopoulos.
\newblock Arm4ch: A methodology for autonomous reality modelling for cultural
  heritage.
\newblock {\em Sensors}, 24(15):4950, 2024.

\bibitem{batinovic2022shadowcasting}
Ana Batinovic, Antun Ivanovic, Tamara Petrovic, and Stjepan Bogdan.
\newblock A shadowcasting-based next-best-view planner for autonomous 3d
  exploration.
\newblock {\em IEEE Robotics and Automation Letters}, 7(2):2969--2976, 2022.

\bibitem{tarbox1995planning}
Glenn~H Tarbox and Susan~N Gottschlich.
\newblock Planning for complete sensor coverage in inspection.
\newblock {\em Computer vision and image understanding}, 61(1):84--111, 1995.

\bibitem{lee2024autonomous}
Inhwan~Dennis Lee, Ji~Hyun Seo, and Byounghyun Yoo.
\newblock Autonomous view planning methods for 3d scanning.
\newblock {\em Automation in Construction}, 160:105291, 2024.

\bibitem{kriegel2015efficient}
Simon Kriegel, Christian Rink, Tim Bodenm{\"u}ller, and Michael Suppa.
\newblock Efficient next-best-scan planning for autonomous 3d surface
  reconstruction of unknown objects.
\newblock {\em Journal of Real-Time Image Processing}, 10:611--631, 2015.

\bibitem{morreale2019predicting}
Luca Morreale, Andrea Romanoni, and Matteo Matteucci.
\newblock Predicting the next best view for 3d mesh refinement.
\newblock In {\em Intelligent Autonomous Systems 15: Proceedings of the 15th
  International Conference IAS-15}, pages 760--772. Springer, 2019.

\bibitem{aravecchia2023next}
St{\'e}phanie Aravecchia, Antoine Richard, Marianne Clausel, and C{\'e}dric
  Pradalier.
\newblock Next-best-view selection from observation viewpoint statistics.
\newblock In {\em 2023 IEEE/RSJ International Conference on Intelligent Robots
  and Systems (IROS)}, pages 10505--10510. IEEE, 2023.

\bibitem{yu2024semantic}
Xiaotong Yu and Chang~Wen Chen.
\newblock Semantic-aware next-best-view for multi-dofs mobile system in
  search-and-acquisition based visual perception.
\newblock In {\em Proceedings of the 32nd ACM International Conference on
  Multimedia}, pages 3713--3721, 2024.

\bibitem{naazare2022online}
Menaka Naazare, Francisco~Garcia Rosas, and Dirk Schulz.
\newblock Online next-best-view planner for 3d-exploration and inspection with
  a mobile manipulator robot.
\newblock {\em IEEE Robotics and Automation Letters}, 7(2):3779--3786, 2022.

\bibitem{hornung2013octomap}
Armin Hornung, Kai~M Wurm, Maren Bennewitz, Cyrill Stachniss, and Wolfram
  Burgard.
\newblock Octomap: An efficient probabilistic 3d mapping framework based on
  octrees.
\newblock {\em Autonomous robots}, 34:189--206, 2013.

\bibitem{border2024surface}
Rowan Border and Jonathan~D Gammell.
\newblock The surface edge explorer (see): A measurement-direct approach to
  next best view planning.
\newblock {\em The International Journal of Robotics Research}, page
  02783649241230098, 2024.

\bibitem{border2018surface}
Rowan Border, Jonathan~D Gammell, and Paul Newman.
\newblock Surface edge explorer (see): Planning next best views directly from
  3d observations.
\newblock In {\em 2018 IEEE International Conference on Robotics and Automation
  (ICRA)}, pages 6116--6123. IEEE, 2018.

\bibitem{border2020proactive}
Rowan Border and Jonathan~D Gammell.
\newblock Proactive estimation of occlusions and scene coverage for planning
  next best views in an unstructured representation.
\newblock In {\em 2020 IEEE/RSJ International Conference on Intelligent Robots
  and Systems (IROS)}, pages 4219--4226. IEEE, 2020.

\bibitem{dhami2023pred}
Harnaik Dhami, Vishnu~D Sharma, and Pratap Tokekar.
\newblock Pred-nbv: Prediction-guided next-best-view planning for 3d object
  reconstruction.
\newblock In {\em 2023 IEEE/RSJ International Conference on Intelligent Robots
  and Systems (IROS)}, pages 7149--7154. IEEE, 2023.

\bibitem{zeng2020pc}
Rui Zeng, Wang Zhao, and Yong-Jin Liu.
\newblock Pc-nbv: A point cloud based deep network for efficient next best view
  planning.
\newblock In {\em 2020 IEEE/RSJ International Conference on Intelligent Robots
  and Systems (IROS)}, pages 7050--7057. IEEE, 2020.

\bibitem{mendoza2020supervised}
Miguel Mendoza, J~Irving Vasquez-Gomez, Hind Taud, L~Enrique Sucar, and
  Carolina Reta.
\newblock Supervised learning of the next-best-view for 3d object
  reconstruction.
\newblock {\em Pattern Recognition Letters}, 133:224--231, 2020.

\bibitem{chen2024gennbv}
Xiao Chen, Quanyi Li, Tai Wang, Tianfan Xue, and Jiangmiao Pang.
\newblock Gennbv: Generalizable next-best-view policy for active 3d
  reconstruction.
\newblock In {\em Proceedings of the IEEE/CVF Conference on Computer Vision and
  Pattern Recognition}, pages 16436--16445, 2024.

\bibitem{wang2024rl}
Tao Wang, Weibin Xi, Yong Cheng, Hao Han, and Yang Yang.
\newblock Rl-nbv: A deep reinforcement learning based next-best-view method for
  unknown object reconstruction.
\newblock {\em Pattern Recognition Letters}, 2024.

\bibitem{pito1999solution}
Richard Pito.
\newblock A solution to the next best view problem for automated surface
  acquisition.
\newblock {\em IEEE Transactions on pattern analysis and machine intelligence},
  21(10):1016--1030, 1999.

\bibitem{gazani2023bag}
Sara~Hatami Gazani, Matthew Tucsok, Iraj Mantegh, and Homayoun Najjaran.
\newblock Bag of views: An appearance-based approach to next-best-view planning
  for 3d reconstruction.
\newblock {\em IEEE Robotics and Automation Letters}, 9(1):295--302, 2023.

\bibitem{kriegel2011surface}
Simon Kriegel, Tim Bodenm{\"u}ller, Michael Suppa, and Gerd Hirzinger.
\newblock A surface-based next-best-view approach for automated 3d model
  completion of unknown objects.
\newblock In {\em 2011 IEEE International Conference on Robotics and
  Automation}, pages 4869--4874. IEEE, 2011.

\bibitem{bendels2006detecting}
Gerhard~H Bendels, Ruwen Schnabel, and Rheinhard Klein.
\newblock Detecting holes in point set surfaces.
\newblock 2006.

\bibitem{coumans2021}
Erwin Coumans and Yunfei Bai.
\newblock Pybullet, a python module for physics simulation for games, robotics
  and machine learning.
\newblock \url{http://pybullet.org}, 2016--2021.

\bibitem{shapenet2015}
Angel~X. Chang, Thomas Funkhouser, Leonidas Guibas, Pat Hanrahan, Qixing Huang,
  Zimo Li, Silvio Savarese, Manolis Savva, Shuran Song, Hao Su, Jianxiong Xiao,
  Li~Yi, and Fisher Yu.
\newblock {ShapeNet: An Information-Rich 3D Model Repository}.
\newblock Technical Report arXiv:1512.03012 [cs.GR], Stanford University ---
  Princeton University --- Toyota Technological Institute at Chicago, 2015.

\bibitem{wu20153d}
Zhirong Wu, Shuran Song, Aditya Khosla, Fisher Yu, Linguang Zhang, Xiaoou Tang,
  and Jianxiong Xiao.
\newblock 3d shapenets: A deep representation for volumetric shapes.
\newblock In {\em Proceedings of the IEEE conference on computer vision and
  pattern recognition}, pages 1912--1920, 2015.

\bibitem{yuksel2015sample}
Cem Yuksel.
\newblock Sample elimination for generating poisson disk sample sets.
\newblock In {\em Computer Graphics Forum}, volume~34, pages 25--32. Wiley
  Online Library, 2015.

\end{thebibliography}

\end{document}